\documentclass[letterpaper, 10 pt, journal, twoside]{IEEEtran}
\IEEEoverridecommandlockouts
\usepackage[pdftex]{graphicx}
\usepackage{fancyhdr}
\usepackage[noend]{algpseudocode}
\usepackage{amsmath}
\usepackage{amssymb}
\usepackage{xcolor}
\usepackage{graphicx}
\usepackage{ulem}
\usepackage{multirow}
\usepackage{url}
\usepackage[linkcolor=black,citecolor=black,urlcolor=black,colorlinks=true]{hyperref}
\usepackage{threeparttable}
\usepackage{makecell}
\usepackage{cite}
\usepackage[linesnumbered,boxed,ruled,commentsnumbered]{algorithm2e}
\graphicspath{{figures/}}
\DeclareGraphicsExtensions{.pdf,.png,.jpg,.eps}
\usepackage[font=small,skip=5pt]{caption}

\begin{document}
%

\title{\LARGE \bf Efficient and Probabilistic Adaptive Voxel Mapping for Accurate Online \textcolor{black}{LiDAR Odometry}}

\author{Chongjian Yuan$^{12}$,
        Wei Xu$^{1}$, Xiyuan Liu$^{1}$,
        Xiaoping Hong$^{2}$,
        and Fu Zhang,$^{1}$

\thanks{$^{1}$C. Yuan, W. Xu, X. Liu and F. Zhang are with the Department of Mechanical Engineering, The University of Hong Kong, Hong Kong Special Administrative Region, People's Republic of China.
{\tt\footnotesize $\{$ycj1,xuweii,xliuaa$\}$@connect.hku.hk}, {\tt\footnotesize $ $fuzhang$ $@hku.hk}}
\thanks{$^{2}$C. Yuan and X. Hong are with the School of System Design and Intelligent Manufacturing, Southern University of Science and Technology, Shenzhen, People’s Republic of China.{\tt\footnotesize $\{$yuancj2020,hongxp$\}$@sustech.edu.cn}}

}
\maketitle

\begin{abstract}
This paper proposes an efficient and probabilistic adaptive voxel mapping method for \textcolor{black}{LiDAR odometry}. The map is a collection of voxels; each contains one plane feature that enables the probabilistic representation of the environment and accurate registration of a new LiDAR scan. We further analyze the need for coarse-to-fine voxel mapping and then use a novel voxel map organized by a Hash table and octrees to build and update the map efficiently. We apply the proposed voxel map to an iterated extended Kalman filter and construct a maximum a posteriori probability problem for pose estimation. Experiments on the open KITTI dataset show the high accuracy and efficiency of our method compared to other state-of-the-art methods. \textcolor{black}{Experiments on indoor and  unstructured outdoor environments with solid-state LiDAR and non-repetitive scanning LiDAR} further verify the adaptability of our mapping method to different environments and LiDAR scanning patterns (see our attached video\footnote{\url{https://youtu.be/HSwQdXg31WM}}). \textcolor{black}{Our codes and dataset are open-sourced on Github\footnote{\url{https://github.com/hku-mars/VoxelMap}}
}

\end{abstract}


%
\IEEEpeerreviewmaketitle

\vspace{-0.2cm}
\section{Introduction}
\label{sec:intro}
\IEEEPARstart{I}n recent years, with the development of 3D LiDAR technology, especially the emergence of low-cost high-density solid-state LiDARs \cite{liu2020low},  LiDAR(-inertial) odometry has been increasingly used in various applications such as autonomous vehicles \cite{geiger2012we}, UAVs (Unmanned Aerial Vehicles) \cite{xu2020fastlio, okubo2009characterization} and 3D mobile mapping devices \cite{ravi2020pothole}.

A fundamental requirement of LiDAR odometry is a proper representation of the environment (i.e., map) that can be maintained efficiently and can register new points effectively. A predominant mapping method in  existing LiDAR(-inertial) odometry, such as LOAM \cite{zhang2014loam}, Lego-LOAM \cite{shan2018lego}, LINS \cite{qin2020lins}, LIO-SAM \cite{liosam2020shan}, and Lili-OM \cite{liliom2021}, is a point cloud map, which consists of raw or selected edge or plane points. When registering a new scan, each point in the scan is registered to a small plane fitted by a few nearest points in the map by iterated closet point (ICP) \cite{sharp2002icp} or its variants (e.g., generalized ICP \cite{gicp}). This line of work consummates in FAST-LIO2\cite{fastlio2}, which develops an incremental kd-tree structure to organize the point cloud map efficiently. 

As a direct form of LiDAR measurements, a point cloud map is simple to implement. However, a major drawback of the point cloud map is difficult to consider map uncertainty caused by LiDAR measurement noise. In ICP or its variants, the plane fitted by the nearest map points is usually regarded as the true deterministic plane that a new point lies on, while it should really be probabilistic due to measurements noises in the map points. While incorporating these noises in the fitted plane (and the subsequent point registration) is possible, it is less reliable (since the nearest points are usually few in order to avoid spurious planes) and time-consuming (since the uncertainty has to be recomputed if any nearest point is changed during the registration). 

Consideration of the map uncertainty requires explicit parameterization of salient features (e.g., plane) in the environment, tracking these features across different LiDAR scans and estimating these feature parameters and their uncertainties. These tasks are challenging due to the various possible environments which contain features of different sizes. The problem is further complicated by the coarse-to-fine phenomenon widely in LiDAR measurements. For example, points in a scan or map usually have drastically different densities in the space due to varying LiDAR resolution, scanning types, and considerably uneven distribution of points in the space. The point density also varies over time due to the sequentially scanning nature of LiDAR sensors.

\textcolor{black}{To address the above challenges, this paper proposes a novel online adaptive voxel mapping method, which constructs voxels of different sizes to adapt to the variation in environment structures and point density. A voxel tracks one salient feature (we use plane in this paper). The feature parameters and uncertainties are repeatedly estimated within the voxel based on received points. The parameterized feature allows us to consider the uncertainty of both the current point and the map accurately. Concretely, the contributions of this paper include:}
\begin{enumerate}
    \item  An adaptive-size and coarse-to-fine voxel construction method, which is adaptable to environments of varying structures and robust to the sparsity and irregularity of LiDAR point cloud. The adaptive voxel map is organized in an octree-hash data structure to increase the efficiency of voxel construction, updates, and inquiries.
    \item A true probabilistic map representation, where each feature (i.e., plane) contained in the voxel map accurately considers uncertainties arising from both point measurement noise and pose estimation error.
    \item \textcolor{black}{An open-source implementation of the proposed mapping method in a LiDAR\textcolor{black}{(-inertial) }odometry system and full validation of the design on real-world datasets of various environments (structured and unstructured) and LiDARs (multi-spinning LiDARs and non-conventional solid-state LiDARs). In particular, the superiority of our method over other state-of-art methods on KITTI dataset is demonstrated. }
    
\end{enumerate}





\vspace{-0.2cm}
\section{Related Works}

\textcolor{black}{On the one hand, the plane feature used in our voxel map is most similar to the surfel used in Suma\cite{suma2018rss} and Suma++\cite{chen2019suma++}, which also register new points by optimizing the point to plane residuals and explicitly maintain and estimate the surfel in the map. The surfel in \cite{suma2018rss, chen2019suma++} is a very small planar feature, which is also the case of our plane features in cluttered environments. However, due to the adaptive sizes in our voxel maps, our plane features could be very large in structured environments.  This adaptive strategy makes our method suitable for environments with varying structures \cite{wisth2021unified} while ensuring high computation efficiency. Another difference between our plane features with the surfels in \cite{suma2018rss} and \cite{chen2019suma++} lies in the surfel/plane association. In \cite{suma2018rss, chen2019suma++}, both surfels in the map and new LiDAR points are projected (i.e., rendered) to a virtual image plane to perform point-to-plane association. Such a surfel rendering process is very time-consuming and typically requires GPU acceleration for online applications. In contrast, our plane association is made by efficient Hash table inquiry. }

\textcolor{black}{The computation issue in the surfel rendering could be mitigated by modeling the surfel map as a Gaussian mixture model and performing surfel association by Expectation Maximization, as shown in MRSLaserMap\cite{marslasermap} and \cite{quenzel2021real}. The works in \cite{marslasermap, quenzel2021real} also adopted a multi-resolution surfel map, where the valid surfels are merged from the finest resolution to coarser resolution. This fine-to-coarse map method originates from MRSMap\cite{stuckler2014multi}, \textcolor{black}{which was inspired from the radial nature of the LiDAR scanner, }
and is usually sensitive to the sparsity and irregularity of the LiDAR point cloud. In contrast, our method constructs voxel maps in a coarse-to-fine manner and is more robust to LiDAR point cloud’s irregularity and adaptable to different environments.}

\textcolor{black}{Instead of deterministic surfels, the works in \cite{park2017probabilistic, elastic-continuous-tro} incorporate LiDAR measurements noises into surfels, leading to a probabilistic point-to-surfel registration similar to our probabilistic plane representation. However, \cite{park2017probabilistic, elastic-continuous-tro} only considered the LiDAR measurement noises but ignored the uncertainties caused by errors in LiDAR pose estimation. In contrast, our plane uncertainty considers both uncertainties. Notice that the pose estimate is used to transform LiDAR points measured in the local body frame to the world frame where the surfel/plane is estimated, hence the pose estimation error will contribute to the surfel/plane uncertainty. }


On the other hand, our voxel map is similar to the normal distribution transformation (NDT) \cite{magnusson2007scan}, where the map consists of many voxels (of different sizes, e.g., \cite{2013ndtom}). In each voxel, a 3D Gaussian distribution is fitted from map points lying in the voxel and is used to register a new point by maximizing the likelihood of the point being taken from the distribution. LiTAMIN2\cite{litamin2} further improves the accuracy of NDT by incorporating ICP with NDT. When compared with these NDT methods, our method explicitly parameterizes and estimates a plane in a voxel. Such explicit plane parameterization allows performing direct point-to-plane registration, which is more sound and often accurate than NDT where the 3D Gaussian distribution over constrains the new point by penalizing all three directions (instead of the plane normal only). The explicit plane parameterization also allows incorporating noises in both LiDAR point measurements and pose estimation, which provides further accuracy improvements. 

\textcolor{black}{In terms of data structure, our proposed octo-hash data structure and coarse-to-fine mapping method are similar to OGN \cite{tatarchenko2017octree} and HSP \cite{hane2017hierarchical}. OGN \cite{tatarchenko2017octree} introduces an octo-hash representation into a deep convolution decoder, and achieves computation- and memory-efficient generation of volumetric 3D output.
 HSP \cite{hane2017hierarchical}  uses a coarse-to-fine prediction on a hierarchical surface for a more accurate 3D object reconstruction. While good results of 3D object output and object reconstruction from OGN \cite{tatarchenko2017octree} and HSP \cite{hane2017hierarchical}
  are presented, we integrate the efficient octo-hash data structure and coarse-to-fine plane representation into a new type of map, and apply the map to the LiDAR odometry problem, enhancing the adaptability, efficiency and accuracy of LiDAR odometry.}

\section{Methodology}
\subsection{Probabilistic Plane Representation\label{sec:true-probalistic}}
\textcolor{black}{Our voxel map contains one probabilistic feature in each voxel. Without loss of generality, we use plane features due to their vast availability in environments and present the uncertainty  model of a plane feature in this section. }

\textcolor{black}{Since a plane feature is estimated from its associated points, any noises in the points will contribute to the plane estimation uncertainty. We first investigate the point noises in Section \ref{point variance analysis} and then derive how the noises contribute to the plane uncertainty in Section \ref{surfel uncertainty model}. Notice that since the voxel map (hence the contained plane features) is represented in the world frame, the point noises should also be investigated in the world frame, which leads to two possible noise sources: one is the raw point measurement noises, which is relative to the local LiDAR body frame, and the other is the error in LiDAR pose estimation that projects the local LiDAR points to the world frame. } 

\subsubsection{Uncertainty of point ${}^W \! \mathbf p_i$ \label{point variance analysis}}
\
\par
\begin{figure}[t]
    \begin{center}
    \setlength{\abovecaptionskip}{-0.2cm}
        {\includegraphics[width=0.9\columnwidth]{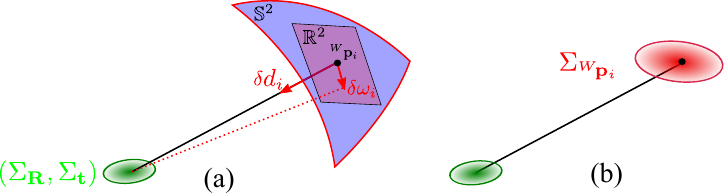}}
    \end{center}
    \caption{The uncertainty model of the registered point: (a) the LiDAR measurement uncertainty model, including bearing direction and depth measurement uncertainty; (b) point covariance $ \boldsymbol{\Sigma}_{{}^W \! \mathbf p_i}$ computed from (\ref{eq:point_cov}).}
    \label{fig:point_cov}
\end{figure}

According to the analysis of the measurement noises for LiDAR sensors in \cite{pixel-level}, the uncertainty of a LiDAR point in the local LiDAR frame consists of two parts, the ranging uncertainty and the bearing direction uncertainty (see Fig. \ref{fig:point_cov} (a)). Let $\boldsymbol{\omega}_i \in \mathbb{S}^2$ be the measured bearing direction,  $\boldsymbol{\delta}_{\boldsymbol{\omega}_i} \sim \mathcal{N}(\mathbf 0_{2\times 1}, \boldsymbol{\Sigma}_{\boldsymbol{\omega}_i})$ be the bearing direction noise in the tangent plane of $\boldsymbol{\omega}_i$, $d_i$ be the depth measurement and $\delta_{d_i} \sim \mathcal{N}(0, {\Sigma}_{{d}_i}) $ be the ranging noise. Then the noise $\boldsymbol{\delta}_{ {}^L \! \mathbf p_i}$ of the measured point ${}^L \mathbf p_i $ and  its covariance $ \boldsymbol{ \Sigma}_{{}^L \! \mathbf p_i}$ is
\begin{equation}\footnotesize
    \begin{split}
        \boldsymbol{\delta}_{ {}^L \! \mathbf p_i}&= \underbrace{\begin{bmatrix}
        \boldsymbol{\omega}_i & - d_i \lfloor \boldsymbol{\omega}_i \rfloor_{\wedge} \mathbf N(\boldsymbol{\omega}_i)
    \end{bmatrix}}_{\mathbf A_i }
    \begin{bmatrix}
     \delta_{d_i} \\
     \boldsymbol{\delta}_{\boldsymbol{\omega}_i}
    \end{bmatrix} \sim \mathcal{N}(\mathbf 0, \boldsymbol{\Sigma}_{{}^L \! \mathbf p_i}), \\
    \boldsymbol{\Sigma}_{{}^L \! \mathbf p_i} &= \mathbf A_i 
    \begin{bmatrix}
    \Sigma_{d_i} & \mathbf 0_{1 \times 2}  \\
    \mathbf 0_{2 \times 1} & \boldsymbol{\Sigma}_{\boldsymbol{\omega}_i}
    \end{bmatrix}\mathbf A_i^T.
        \end{split}
    \label{eq:meas_noise_cov}
\end{equation}
where $\mathbf N(\boldsymbol{\omega}_i) = \begin{bmatrix}
\mathbf N_1 & \mathbf N_2
\end{bmatrix} \in \mathbb{R}^{3 \times 2} $ is an orthonormal basis of the tangent plane at $\boldsymbol{\omega}_i$, $\lfloor  \  \rfloor_{\wedge} $ denotes the skew-symmetric matrix mapping the cross product. Detailed derivation of equation (\ref{eq:meas_noise_cov}) can be found in \cite{pixel-level}.

Considering that the LiDAR point ${}^L  \mathbf p_i$ will be further projected to the world frame through the estimated pose ${}^W_L \bar{\mathbf T} = ({}^W_L {\mathbf R} , {}^W_L {\mathbf t} ) \in SE(3)$, with estimation uncertainty $\left( \Sigma_{\mathbf R}, \Sigma_{\mathbf t} \right)$ as shown in Fig.\ref{fig:point_cov} (a), by the following rigid transformation
\begin{equation}
    {}^W \! \mathbf p_i = {}^W_L {\mathbf R} {}^L \mathbf p_i + {}^W_L {\mathbf t}
\end{equation}
Therefore, the uncertainty of the LiDAR point ${}^W \mathbf p_i$ is hence
\begin{equation} \label{eq:point_cov}
     \boldsymbol{\Sigma}_{{}^W \! \mathbf p_i} = {}^W_L {\mathbf R} \boldsymbol{\Sigma}_{{}^L \! \mathbf p_i} {}^W_L {\mathbf R}^T +{}^W_L {\mathbf R} \lfloor {}^L \mathbf p_i \rfloor_{\wedge} \boldsymbol{\Sigma}_{\mathbf R} \lfloor {}^L \mathbf p_i \rfloor_{\wedge} ^T {}^W_L {\mathbf R}^T+ \boldsymbol{\Sigma}_{\mathbf t}
\end{equation}
where $\boldsymbol{\Sigma}_{\mathbf R}$ is the uncertainty of the ${}^W_L {\mathbf R}$  in the tangent space and $\boldsymbol{\Sigma}_\mathbf t$ is the uncertainty of the ${}^W_L {\mathbf t}$.
Considering the uncertainty due to LiDAR measurement and pose estimation, the covariance of point cloud at different positions has a great difference: points at closer distances have their noises dominated by the ranging noise while at further distances dominated by the bearing noise. The uncertainty analysis of a single LiDAR point is  also the basis of the uncertainty model of a plane feature.
\subsubsection{Plane Uncertainty Modelling\label{surfel uncertainty model}}
\
\par

\begin{figure}[t]
    \vspace{-0.2cm}
    \begin{center}
        {\includegraphics[width=0.7\columnwidth]{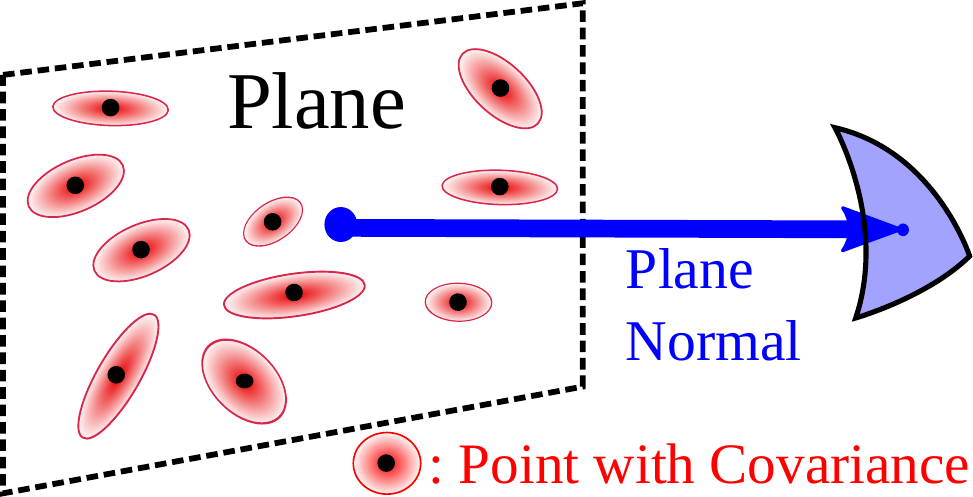}}
    \end{center}
    \caption{\label{fig:surfel normal}The Uncertainty model of the plane normal.}
    \vspace{-0.4cm}
\end{figure}

Let a plane feature consist of a group of LiDAR points ${}^W \! \mathbf p_i$ $(i=1,...,N)$, each has an uncertainty ${}^W \!  \boldsymbol{\Sigma}_{\mathbf p_i}$ due to the measurement noise and pose estimation error as shown in (\ref{eq:point_cov}). Denote the points covariance matrix be A:
\begin{equation}\label{mean_cov}
\footnotesize
    \bar{\mathbf p}=\frac{1}{N}\sum_{i=1}^N {}^W \! \mathbf p_i; \quad \mathbf A=\frac{1}{N}\textcolor{black}{\sum_{i=1}^N} \left({}^W \! \mathbf p_i -\bar{\mathbf p} \right) \left({}^W \! \mathbf p_i -\bar{\mathbf p} \right)^T;
\end{equation}
Then, the plane can be represented by its normal vector $\mathbf n$, which is the eigenvector associated with the minimum eigenvalue of $\mathbf A$, and the point $\mathbf q = \bar{\mathbf p}$, which lies in this plane. As both $\mathbf A$ and $\bar{\mathbf p}$ are dependent on ${}^W \! \mathbf p_i$, we can denote the plane parameters $(\mathbf n, \mathbf q)$ as functions of ${}^W \! \mathbf p_i$ shown below: 
\begin{equation}
 \label{nq_func}
[\mathbf n, \mathbf q]^T= \mathbf f({}^W \! \mathbf p_1,{}^W \! \mathbf p_2,...,{}^W \! \mathbf p_N).
\end{equation}
Based on the uncertainty analysis in \ref{point variance analysis}, each LiDAR point ${}^W \! \mathbf p_i$ has a noise $\boldsymbol{\delta}_{{}^W \! \mathbf p_i} \sim \mathcal{N}(\mathbf 0_{3\times 1},\boldsymbol{\Sigma}_{{}^W \! \mathbf p_i}) $. Therefore, the ground-truth normal vector $\mathbf n^{gt}$ and ground-truth position $\mathbf q^{gt}$ \textcolor{black}{(the real value in the real world that is to be estimated)} are
\begin{equation}
\begin{aligned}
\footnotesize
	\left[ \mathbf n^{gt},\mathbf q^{gt} \right]^T &= \mathbf f({}^W \! \mathbf p_1 + \boldsymbol{\delta}_{{}^W \! \mathbf p_1},{}^W \! \mathbf p_2 + \boldsymbol{\delta}_{{}^W \! \mathbf p_1},...,{}^W \! \mathbf p_N + \boldsymbol{\delta}_{{}^W \! \mathbf p_{\! N}})\\
    & \approx     
\left[ \mathbf n,\mathbf q \right]^T + \sum_{i=1}^N \frac{\partial \mathbf f}{\partial {}^W \! \mathbf p_i} \boldsymbol{\delta}_{{}^W \! \mathbf p_i}
\end{aligned}
\end{equation}
Here, $\frac{\partial \mathbf f}{\partial {}^W \! \mathbf p_i}=[\frac{\partial \mathbf n}{\partial {}^W \! \mathbf p_i},\frac{\partial \mathbf q}{\partial {}^W \! \mathbf p_i}]^T$. Assume  $\mathbf A$ has the eigenvector matrix $\mathbf U$, the minimum eigenvalue $\lambda_3$ and the corresponding eigenvector $\mathbf u_3$, referring to \cite{liu2021balm}, we can take the derivative of the $\mathbf n$ and $\mathbf q$ with respect to each point ${}^W \! \mathbf p_i$ as below:
\begin{equation}
\footnotesize \notag
    \frac{\partial \mathbf n}{\partial {}^W \! \mathbf{p}_i}\!=\!
    \mathbf U\!\!
    \begin{bmatrix}
    \mathbf F_{1,3}^{\mathbf p_i}\\
    \mathbf F_{2,3}^{\mathbf p_i}\\
    \mathbf F_{3,3}^{\mathbf p_i}
    \end{bmatrix}\! \! ,
\label{eq6}
\mathbf F_{m,3}^{\mathbf p_i} \! = \! \left\{
\begin{aligned}
\frac{({}^W \! \mathbf p_i \! - \! \mathbf q)^T}{N(\lambda_3 \! - \! \lambda_m)}(& \mathbf u_m \mathbf n^{\!T} \!\!  + \! \mathbf n \mathbf u_m^{\! T})\!\!  & \! \! ,m \! \neq \! 3,\\
 & \mathbf 0_{1\times3}\!\!  &\! \! ,m\! =\! 3.
\end{aligned}
\right.  
\end{equation}
\begin{equation}  \footnotesize
   \frac{\partial \mathbf q}{\partial {}^W \! \mathbf p_i} = \mathbf{diag} \left(\frac{1}{N},\frac{1}{N},\frac{1}{N} \right)
\end{equation}
Then covariance matrix $\boldsymbol{\Sigma}_{\mathbf n, \mathbf q}$ of $\mathbf n$ and $\mathbf q$ is therefore:
\begin{equation}
\footnotesize
    \label{eq:plane uncertainty model}
    \boldsymbol{\Sigma}_{\mathbf n, \mathbf q} = \sum_{i=1}^N\frac{\partial \mathbf f}{\partial {}^W \! \mathbf p_i} \boldsymbol{\Sigma}_{{}^W \! \mathbf p_i} \frac{\partial \mathbf f}{\partial {}^W \! \mathbf p_i}^T
\end{equation}
It is seen that $\mathbf n$ and $\mathbf q$ are not independent as they are calculated from the same set of noisy points.

\vspace{-0.05cm}
\subsection{Coarse-to-fine and Efficient Voxel Map Construction}


In this section, we first explain the motivation for a coarse-to-fine voxel-based map and then the methodology to build and update the voxel map efficiently.
\subsubsection{Motivation}
\
\par
\begin{figure}[t]
    \setlength{\abovecaptionskip}{-0.05cm}
    \begin{center}
        {\includegraphics[width=0.9\columnwidth]{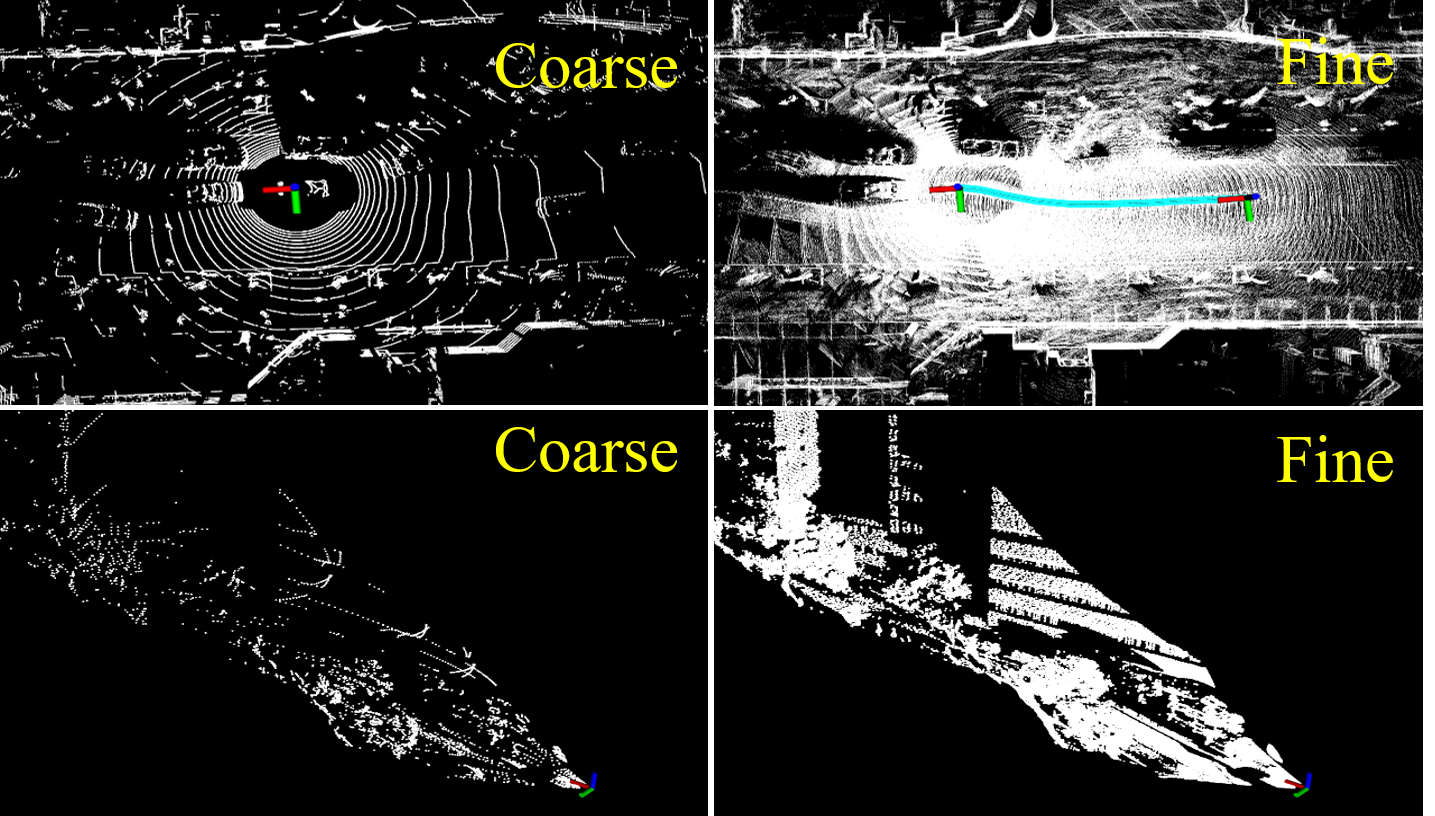}}
    \end{center}
    \caption{\label{fig:coar_to_fine}The coarse-to-fine phenomenons in a spinning LiDAR (top two figures) and a non-repetitive solid-state LiDAR (bottom two figures).}
    \vspace{-0.4cm}
\end{figure}

As shown in Fig. \ref{fig:coar_to_fine}, LiDAR points are typically sampled sequentially,  hence a scan of the point cloud is always accumulated from sparse to dense, especially in outdoor environments where the points are distributed in a larger space. When the point cloud is relatively sparse, common surfel-based fine-to-coarse mapping methods can usually obtain a very small number of planes only, limiting their applications to high-resolution LiDARs and relatively low scan rate (e.g., 10Hz) such that a sufficient number of points could be accumulated. This would, however, require compensating the motion within the accumulated period. To address this issue, we propose a coarse-to-fine voxel mapping method that can build a rough voxel map when the point cloud is sparse and refine the map when more points are received.

\subsubsection{Voxel Map Construction \label{sec:voxel map}}
\
\par
To achieve voxel map construction in a coarse-to-fine manner, we build an adaptive voxel map organized by a Hash table and an octree for each Hash entry. More specifically, we first cut the space (in the global world frame) into voxels, each with the size of the coarse map resolution. Then, for the first LiDAR scan, which defines the world frame, the contained points are distributed into voxels. Voxels populated with points are indexed into a Hash table. Then, for each populated voxel, if all the contained points lie on a plane (the minimum eigenvalue of the point covariance matrix is less than a specified threshold), we store the plane points and calculate the plane parameters $(\mathbf n, \mathbf q)$ as in (\ref{nq_func}) and their uncertainty $\boldsymbol{\Sigma}_{\mathbf n, \mathbf q}$ as in (\ref{eq:plane uncertainty model}); otherwise, the current voxel will break into eight octants and repeat plane checking and voxel cutting in each one until reaching the maximum number of layers. Notice that the voxels have different sizes, each voxel contains one plane feature fitted from the contained LiDAR raw points.

\subsubsection{\textcolor{black}{Voxel} Map Update}
\
\par
\begin{figure}[t]
    \centering
    \includegraphics[width=0.9\linewidth]{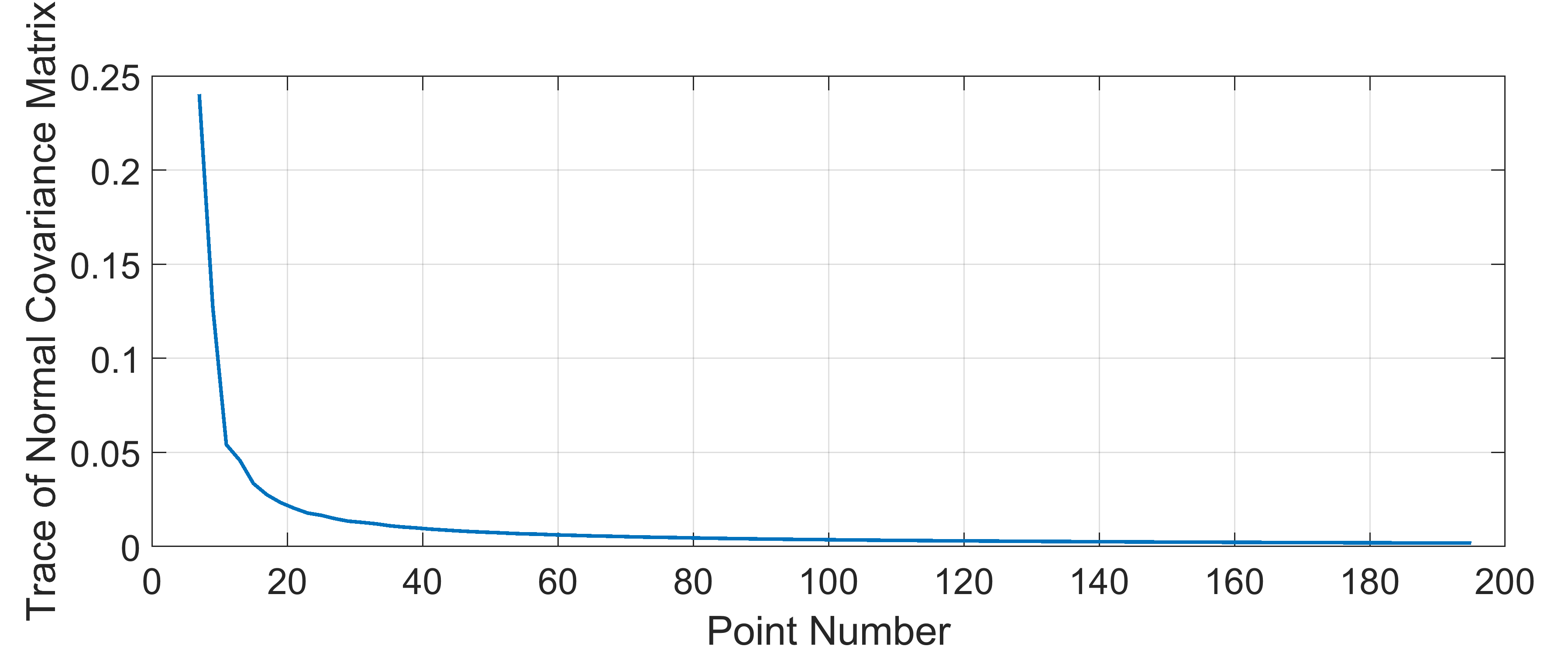}
    \caption{The convergence of the trace of the normal vector covariance matrix with the increase of the number of points.}
    \vspace{-0.4cm}
    \label{fig:trace_point_num}
\end{figure}

For online LiDAR odometry, new scans of LiDAR point cloud are continuously coming and their poses are estimated as in Section \ref{e:state_estimate}. Then the estimated pose is used to register the new points into the global map. When the new points lie in an unpopulated voxel, it will construct the voxel. Otherwise, when the new points are added to an existing voxel, the parameters and the uncertainty of the plane in the voxel should be updated. This will cause a growing processing time with the constant reception of new points. To address this issue, we find that the uncertainty of the plane parameters will quickly converge as illustrated in Fig. \ref{fig:trace_point_num}, where the position of each point carries a Gaussian noise with a zero mean and a variance of $0.1m^2$. It is seen that the uncertainty of the normal vector converges when the number of points reaches 50. After the uncertainty converges, we discard all historical points and retain the estimated plane parameters $(\mathbf n, \mathbf q)$ and covariance $\boldsymbol{\Sigma}_{\mathbf n, \mathbf q}$. Once more new points are coming, we keep only the latest 10 points and calculate the new plane normal vector composed of those 10 points. If the new normal vector and the previously converged normal vector continue to appear a relatively large difference, we assume that this area of the map has changed and needs to be reconstructed as \textcolor{black}{Section \ref{sec:voxel map}}.

\subsection{Point-to-plane Match \label{sec:potion-to-surface}}
This section details how to match points in a new LiDAR scan with the voxel map to construct the constrain for the pose estimation and subsequent point cloud registration. Based on the accurate uncertainty modeling of the points and planes, we could easily implement the point-to-plane scan match.
Given a LiDAR point ${}^W \! \mathbf P_i$ predicted in the world frame with the pose prior, we first find which root voxel (with the coarse map resolution) it lies in by its Hash key. Then, all the contained sub-voxels are polled for a possible match with the point. Specifically, let a sub-voxel contains a plane with normal $\mathbf n_i$ and center $\mathbf q_i$, we calculate the point-to-plane distance:
\begin{equation}
    d_i = \mathbf n_i^T ({}^W \! \mathbf p_i-\mathbf q_i)
    \label{eq:point_to_plane}
\end{equation}
As analyzed above, the normal vector $\mathbf n_i$, the LiDAR point ${}^W \! \mathbf p_i$, and the center $\mathbf q_i$ have uncertainties. Considering all these uncertainties, we obtain:
\begin{equation}
\footnotesize \notag
\begin{aligned}
    d_i&=(\mathbf n^{gt}_i \boxplus  \boldsymbol{\delta}_{\mathbf n_i})^T [({}^W \! \mathbf p^{gt}_i+ \boldsymbol{\delta}_{{}^W \! \mathbf  p_i})-\mathbf q^{gt}_i-\boldsymbol{\delta}_{\mathbf q_i}]\\
    &\approx
    \underbrace{{\mathbf n^{gt}_i}^T({}^W \! \mathbf p^{gt}_i-\mathbf q^{gt}_i)}_{\mathbf 0}\mathbf +
    \underbrace{\mathbf J_{\mathbf n_i} \boldsymbol{\delta}_{\mathbf n_i}+\mathbf J_{ \mathbf q_i} \boldsymbol{\delta}_{ \mathbf q_i}+ \mathbf J_{{}^W \! \mathbf p_i} \boldsymbol{\delta}_{{}^W \! \mathbf p_i} }_{\mathbf w_i}
\end{aligned}
\end{equation}
which implies
\begin{equation}
    d_i \sim \mathcal N(0, \Sigma_{\mathbf w_i}), \quad \Sigma_{\mathbf w_i}= \mathbf J_{\mathbf w_i} \boldsymbol{\Sigma}_{\mathbf n_i, \mathbf q_i, {}^W \! \mathbf p_i}  \mathbf J_{\mathbf w_i}^T,
    \label{eq:distance_sigma}
\end{equation}
where,
\begin{equation}
\begin{aligned}
    &\mathbf J_{\mathbf w_i} = \left [ \mathbf J_{\mathbf n_i}, \mathbf J_{\mathbf q_i},\mathbf J_{{}^W \! \mathbf p_i}\right ] = \left [ ({}^W \! \mathbf p_i - \mathbf q_i)^T,-\mathbf n_i^T, \mathbf n_i^T \right ] \\
    &\boldsymbol{\Sigma}_{\mathbf n_i, \mathbf q_i, {}^W \! \mathbf p_i} = 
    \begin{bmatrix}
    \boldsymbol{\Sigma}_{\mathbf n_i,\mathbf q_i} & 0 \\
    0 & \boldsymbol{\Sigma}_{{}^W \! \mathbf p_i}
    \end{bmatrix}.
\end{aligned}
\end{equation}

That is being said, if the point lies on the candidate plane, its distance $d_i$ should be subject to the distribution in (\ref{eq:distance_sigma}). \textcolor{black}{Therefore, let the standard deviation of the distribution be  $\sigma = \sqrt{\Sigma_{\mathbf w_i}}$}, we can examine whether the measured point-to-plane distance falls within $3\sigma$. If so, it is selected to be an effective match. Besides, if a point matches more than one plane based on the $3\sigma$ criterion, the plane with the highest probability will be matched. If no plane passes the $3 \sigma$ test, the point is discarded to remove the possible false matches caused by voxel quantization.  

\vspace{-0.1cm}
\subsection{State Estimation}\label{e:state_estimate}
We build a LiDAR\textcolor{black}{(-inertial)} odometry system based on an iterated extended Kalman filter similar to FAST-LIO2 \cite{fastlio2}. Assume that we are given a state estimation prior $\widehat{\mathbf x}_k$ with covariance $\widehat{\mathbf P}_k$. \textcolor{black}{The prior can be provided from a constant velocity assumption for LiDAR-only Odometry (experiment A and B) or IMU propagation for LiDAR-inertial Odometry (experiment C).}
This prior will be fused with the point-to-plane distance matched in Section \ref{sec:potion-to-surface} to form a maximum a posteriori (MAP) estimation. Specifically, the $i$-th valid point-to-plane match leads to the observation equation
\begin{equation}
\label{eq:observation equation}
    \begin{aligned}
           \mathbf z_i =  \mathbf h_i(\mathbf x_k) + \mathbf v_i
    \end{aligned}
\end{equation}
where $\mathbf z_i$ is the point-to-plane distance residual $d_i$ in equation (\ref{eq:point_to_plane}), $\mathbf h_i(\mathbf x_k)$ is the observation function and $\mathbf v_i \sim (0,\mathbf R_i)$ is the observation noise. By substituting the state $\mathbf x_k$ (i.e., the sensor pose $\mathbf T_k$) into (\ref{eq:point_to_plane}) and linearize it around the current state update $\bar{\mathbf x}_k$, we obtain: 
\begin{equation}
    \footnotesize \notag
    \begin{aligned}
    \mathbf z_i &= \mathbf h_i (\mathbf x_{k}) + \mathbf v_{i} = \mathbf n_i^T ({}^W \! \mathbf p_i-\mathbf q_i)\\
     &= \! (\mathbf n^{gt}_i \boxplus \boldsymbol{\delta}_{\mathbf n_i})^T [(\mathbf T^{gt}_k \boxplus \boldsymbol{\delta}_{\mathbf T_k})({}^L \mathbf p_i^{gt}+ \boldsymbol{\delta}_{{}^L \! \mathbf  p_i})-\mathbf q^{gt}_i- \boldsymbol{\delta}_{\mathbf q_i}] \\
    &\approx
    \underbrace{{\mathbf n^{\! gt}_i}^{T} \! (\mathbf T^{ gt}_k {}^L \! \mathbf p^{\! gt}_i \! - \! \mathbf q^{gt}_i)}_{\mathbf 0} \! +
    \underbrace{\mathbf J_{\mathbf T_i} \delta_{\mathbf T_k}}_{\mathbf H_i \delta \mathbf x_k} \!+\!
    \underbrace{\mathbf J_{\mathbf n_i} \boldsymbol{\delta}_{\mathbf n_i}\! +\! \mathbf J_{ \mathbf q_i} \boldsymbol{\delta}_{ \mathbf q_i}\! + \! \mathbf J_{{}^L \! \mathbf p_i} \boldsymbol{\delta}_{{}^L \! \mathbf p_i} }_{\mathbf v_i}
    \end{aligned}
\end{equation}
which implies
\begin{equation}
  \mathbf R_i=\mathbf J_{\mathbf v_i} \boldsymbol{\Sigma}_{\mathbf n_i, \mathbf q_i, {}^L \! \mathbf p_i}\mathbf J_{\mathbf v_i}^T, 
\end{equation}
where
\begin{equation}
\notag
\begin{aligned}
       &\mathbf J_{\mathbf v_i} \! = \! [\mathbf J_{\mathbf n_i},\mathbf J_{\mathbf q_i}, \mathbf J_{{}^L \mathbf p_i}] \! = \! [(\bar{\mathbf T}_k {}^L \mathbf p_i-\mathbf q_i)^T,-\mathbf n_i^T,\mathbf n_i^T \bar{\mathbf R}_k] \\
       &\boldsymbol{\Sigma}_{\mathbf n_i, \mathbf q_i, {}^L \! \mathbf p_i} = 
    \begin{bmatrix}
    \boldsymbol{\Sigma}_{\mathbf n_i,\mathbf q_i} & 0 \\
    0 & \boldsymbol{\Sigma}_{{}^L \mathbf p_i}
    \end{bmatrix}.
\end{aligned}
\end{equation}

Finally, combining the state prior with all effective measurements, we can obtain the MAP estimation:
\begin{equation}
\begin{split}~\label{equ:MAP}
    \min_{{\mathbf x}_k} \!\left( \| \mathbf x_k \! \boxminus \! \widehat{\mathbf x}_k \|^2_{ \widehat{\mathbf P}_k} \!+\! \sum\nolimits_{i=1}^{m} \| d_i \! - \!  \mathbf H_i \!\cdot\! ({\mathbf x}_{k} \! \boxminus \! \bar{\mathbf x}_k) \|^2_{\mathbf R_i} \!\right)\!
\end{split}
\end{equation}
where the first part is the state prior and the second part is the measurement observation.

\section{Experiments}
\textcolor{black}{In this section, in order to show the high accuracy, efficiency, and adaptability of our mapping method to different environments and LiDAR types, we conduct experiments on three typical LiDARs in three different environments (urban environment, indoor environment, and unstructured outdoor environment), respectively. The details of the LiDAR we used are shown in Table \ref{tab:lidar_type}. In each experiment, we compare our method with state-of-the-art counterparts. All the experiments are conducted on a desktop computer with Intel i7-10700 @ 2.9 $\rm GHz$ with 16 $\rm{GB}$ RAM and Nvidia GeForce GTX 730 with 2 $\rm{GB}$ RAM (in case a GPU is used). }
\begin{table}[h]
    \setlength\tabcolsep{4pt}
    \footnotesize
    \centering
    \begin{tabular}{c c c c c}
        \hline
        Sensor & Type &Frequency & Scan pattern & FoV \\
        \hline
        HDL-64E S2 & Mechanical & $10$Hz & repetitive & $360^\circ \! \times \! 32^\circ$ \\
        Realsense L515 & Solid-state & $30$Hz & repetitive & $70^\circ \! \times \! 55^\circ$\\
        Livox Avia & Solid-state & $10$Hz & non-repetitive & $70^\circ \! \times \! 77^\circ$\\
        \hline
    \end{tabular}
    \caption{
    \textcolor{black}{Different Types of LiDAR}}
    \label{tab:lidar_type}
    \vspace{-0.4cm}
\end{table}

\subsection{\textcolor{black}{Urban Environment Test}}
\label{ex:kitti}
\textcolor{black}{In this experiment, we evaluate our approach on the odometry datasets of the KITTI Vision Benchmark \cite{geiger2012we}, where the LiDAR data is collected by Velodyne HDL-64E S2 in urban environments  and the in-frame motion has been compensated in advance}. As in \cite{deschaud2018imls}, a vertical angle of 0.22 degree is employed to correct the calibration errors in raw point clouds from the KITTI dataset. 

\begin{table*}[ht]
\scriptsize
\centering
\caption{Accuracy (ATE in meters) Comparison on KITTI Odometry Training Sequences}
\label{tab:ate_result}
\vspace{-0.2cm}
\begin{threeparttable}
\begin{tabular}{c c c c c c c c c c c c c c}
\hline
Approach & 00 & 01 &02 &03 & 04 & 05 & 06 & 07 & 08 & 09 & 10 & Avg. of all scans\\
(Total length [m]) & (3724) &(2453) &(5067) &(560) & (393) & (2205) & (1232) & (694) &(3222) &(1705) &(919) & [deg]/[m]\\
\hline
Ours (full) & \textbf{0.9}/\textbf{2.8} & {1.9}/{7.8} & \textbf{1.7}/\textbf{6.1} & {1.2}/\textbf{0.7} & {0.6}/\textbf{0.3}
& 0.8/\textbf{1.2} & \textbf{0.4}/\textbf{0.4} & {0.7}/0.7 & \textbf{1.1}/\textbf{2.3} & \textbf{1.0}/\textbf{1.9} & 1.0/1.1  & \textbf{1.2}/\textbf{2.9}\\
Ours (w/o adaptive) & 1.1/3.1 & 2.0/8.2 & 1.7/7.1 & 1.2/0.7 & 0.7/0/5 & 0.8/1.3 & 0.5/0.5 & 0.9/0.7 & 1.8/3.1 & 1.1/2.0 & 1.3/1.2 & 1.3/3.4\\
Ours (w/o prob.) & 2.2/3.9 & 2.5/9.5 & 1.9/8.1 & 1.4/0.9 & 0.7/0.4 & 1.0/1.2 & 0.5/0.6 & 1.0/0.6 & 3.0 /6.8 & 1.2/1.9 & 1.2/1.5 & 1.8/4.5\\ 
LiTAMIN2\cite{litamin2} & 1.6/5.8 & 3.5/15.9 & 2.7/10.7 & 2.6/0.8 & 2.3/0.7 & 1.1/2.4 & 1.1/0.9 & 1.0/0.6 & 1.3/2.5 &  1.7/2.1 & 1.2/\textbf{1.0} & 1.8/5.1\\
\textcolor{black}{MULLS\cite{pan2021mulls}} & \textcolor{black}{1.7/6.1} & \textcolor{black}{\textbf{1.0}/\textbf{2.4}} & \textcolor{black}{2.4/10.6} &
\textcolor{black}{\textbf{0.7}/0.7} & \textcolor{black}{\textbf{0.2}/0.9} & \textcolor{black}{1.0/2.4} & \textcolor{black}{0.4/0.6} &
\textcolor{black}{\textbf{0.5}/0.6} &
\textcolor{black}{1.9/4.3} & \textcolor{black}{1.4/2.5} &
\textcolor{black}{\textbf{0.5}/1.1} &
\textcolor{black}{1.5/4.8}\\
Suma \cite{suma2018rss} & 1.0/2.9 & 3.2/13.8 & 2.2/8.4 & 1.5/0.9 & 1.8/0.4 & \textbf{0.7}/1.2 & 0.4/0.4 & 0.7/\textbf{0.5} & 1.5/2.8 & 1.1/2.9 & {0.8}/1.3 & 1.4/3.9
\\FAST-LIO2 \cite{fastlio2}& 2.0/3.8 & 2.5/11.9 & 2.8/12.8 & 1.2/0.8 & 0.7/0.5 & 1.2/2.1  & 1.1/1.2 & 1.2/0.8 & 1.8/3.0 & 1.4/2.0 & 1.1/1.5 & 1.8/4.9 \\
Lego-Loam \cite{shan2018lego}& 2.8/6.3 & 3.8/119.4 & 4.1/14.7 & 4.1/0.9 & 3.3/0.8 & 1.9/2.8 & 1.4/0.8 & 1.5/0.7 & 2.5/3.5 & 2.2/2.1 & 1.9/1.8 & 2.8/11.1\\
\hline
\vspace{-0.4cm}
\end{tabular}
\end{threeparttable}
\end{table*}

\begin{figure*}[ht]
    \vspace{-0.1cm}
    \centering
    \includegraphics[width=0.9\linewidth]{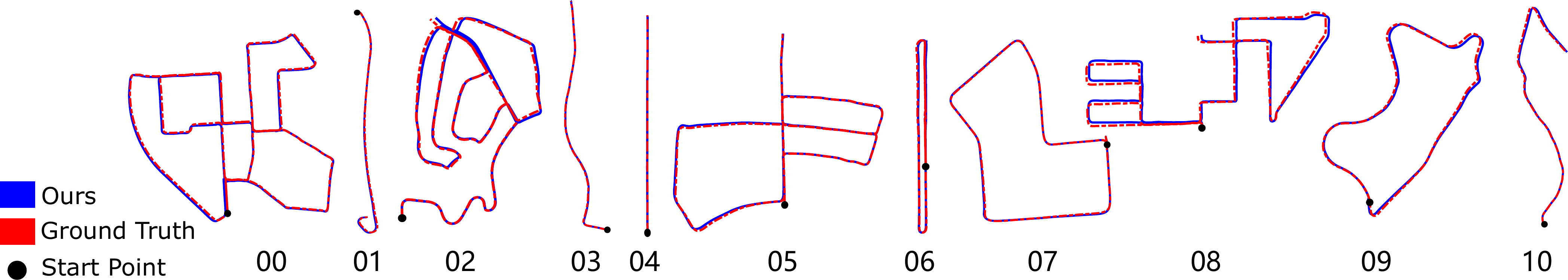}
    \caption{Trajectories of sequence 00 to 10. The blue trajectories are the camera paths obtained from our method and the red trajectories are the gournd truth. The black dot represents the starting point.}
    \vspace{-0.5cm}
    \label{fig:kitti_trajectory}
\end{figure*}

{\bf Comparison with state-of-the-art.}
\label{sec:kitti_experiment}
We compare our method with FAST-LIO2 \cite{fastlio2}, a point-cloud map-based LiDAR-inertial odometry approach, \textcolor{black}{MULLS\cite{pan2021mulls}, a multi-feature-based LO system,} \textcolor{black}{LiTAMIN2\cite{litamin2}, an improved ICP-NDT hybrid method, Suma \cite{suma2018rss}, a surfel-based LiDAR SLAM approach, and Lego-Loam \cite{shan2018lego}, a ground-optimized LiDAR odometry and mapping system.} 

For the implementation, we run our algorithm on all sequences with the same parameters, where the maximum voxel size is 3m and the maximum octo-tree layer is 3, leading to a minimum voxel size of 0.375m. The full version of the algorithm is denoted as \textit{Ours (full)}. To verify the effectiveness of the adaptive voxel module and features probabilistic module, we use a fixed size voxel (i.e., 2m) (termed as \textit{Ours (w/o adaptive))} and a zero plane uncertainty in (\ref{eq:plane uncertainty model}) (termed as \textit{Ours (w/o prob.))}, respectively, while retaining all other modules with the same parameter values. For FAST-LIO2, it is designed for LiDAR-inertial fusion, where IMU propagation is served as a motion prior. Since the KITTI dataset does not have IMU data, we use a constant velocity model similar to ours (see (\ref{equ:MAP})) to serve as the motion prior in FAST-LIO2 for a fair comparison. The other parameters of FAST-LIO2 are their default values recommended for multi-line spinning LiDARs in outdoor environments. 
\textcolor{black}{For MULLS, we conduct the experiments using the open-sourced implementation with the default parameters. }\textcolor{black}{For LiTAMIN2, we directly use the results presented in the original paper\cite{litamin2}. The work in LiTAMIN2\cite{litamin2} further benchmarked their results with Suma \cite{suma2018rss} and Lego-Loam \cite{shan2018lego}, both with loop closure and without loop closure. We directly draw these results and compare them with ours. Since our implementation is a LiDAR odometry without any loop closure, for the sake of fair comparison, we only compare with the results where loop closure for \textcolor{black}{MULLS}, LiTAMIN2, Suma, and Lego-Loam are off.}
\
\par
\textcolor{black}{{\bf Results and Discussion.} Table \ref{tab:ate_result} shows the absolute trajectory error (ATE)\cite{ATE} of all methods, where the results for LiTAMIN2, Suma, and Lego-Loam are directly drawn from \cite{litamin2}. As can be seen, our method (\textit{Ours (full)}) has the best overall accuracy than all other algorithms, especially in long-term sequences like 00, 02 and 08.
 To further show the high accuracy of our method, we plot the estimated LiDAR trajectories of all sequences and their ground-truth in Fig. \ref{fig:kitti_trajectory}. It is seen that all of our global trajectories are very close to the ground truth, verifying the high accuracy of our method. Notice that all variants of our methods use the same parameters across all sequences. }

\textcolor{black}{The improvement of our is mainly attributed to the adaptive voxelization and probabilistic plane representation of the map. The effectiveness of the adaptive voxelization can be seen from the performance degradation from \textit{Ours (full)} to \textit{Ours (w/o adaptive)}. With the adaptive voxelization, our method is able to capture plane features of different sizes (both large-size and finer-level small planes) in the environment. The effectiveness of the probabilistic plane representation can be seen from the performance degradation from \textit{Ours (full)} to \textit{Ours (w/o prob.)}. Without the probabilistic plane representation, the performance of our method is at the same level of FAST-LIO2, which is reasonable because FAST-LIO2 also uses a deterministic plane representation and the same IKFoM framework \cite{ikfom2021he}. Compared to the adaptive voxelization, the probabilistic plane representation plays a much more significant role in the accuracy since the performance degradation by turning off the module is much larger. Finally, it is also interesting to see that our method with only probabilistic plane representation \textit{Ours (w/o adaptive)} can even exceed all comparison methods in most sequences.}

\begin{figure}[h]
    \vspace{-0.1cm}
    \centering
    \includegraphics[width=0.95\linewidth]{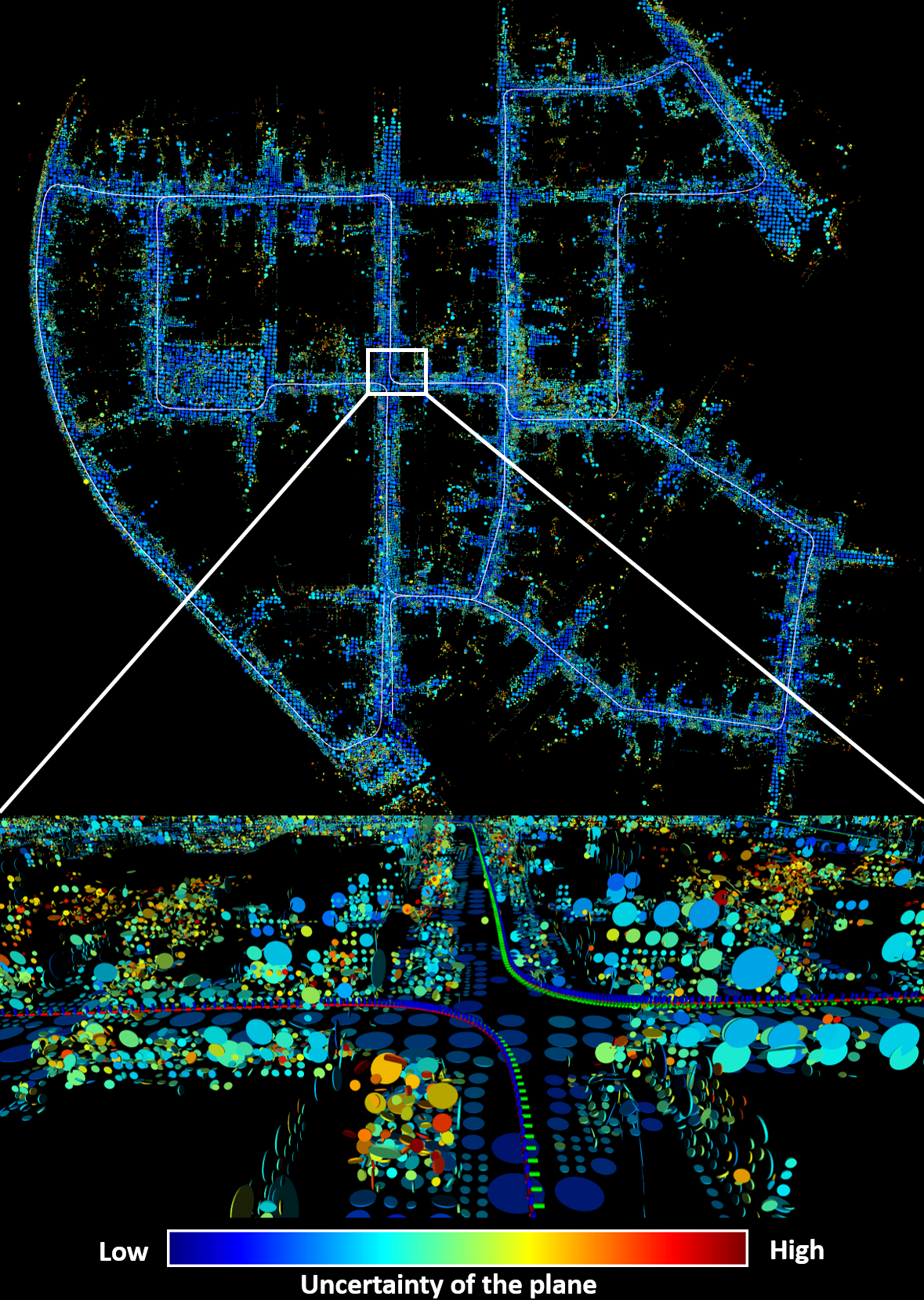}
    \caption{Illustration of plane features in our adaptive voxel maps on KITTI training sequence 00. The plane feature is colored by its uncertainty(the trace of the plane covariance matrix in equatioin (\ref{eq:plane uncertainty model})). The plane is displayed as a disk with radius determined by the maximum eigen value of the points covariance matrix $\mathbf A$ defined in equation (\ref{mean_cov}).}
    \label{fig:kitti_mapping}
    \vspace{-0.5cm}
\end{figure}

\textcolor{black}{Fig. \ref{fig:kitti_mapping} illustrates the plane features estimated in our voxel maps. As can be seen, our voxel map contains plane features of different sizes: planes on the ground and walls are usually large while those on cluttered objects (e.g., moving cars, roadside objects) and their nearby regions are small. Plane features contained in the voxel map are also colored by their respective uncertainty (the trace of the plane covariance matrix (\ref{eq:plane uncertainty model}) in Fig. \ref{fig:kitti_mapping}. As can be seen, plane features far away from the road have much higher uncertainty than that nearby. This is caused by the uneven LiDAR point measurements, where nearby areas have many more measurements than far areas due to the repeat observation. We also notice that plane features in cluttered areas have higher uncertainties than that in structured areas (e.g., ground, wall) and that plane features on moving objects (e.g., cars) have higher uncertainties than static ones. All these phenomena are as expected. }
\
\par
{\bf Runtime.} Besides accuracy, we also evaluate the computation efficiency of our method against others. Since the LiTAMIN2 \cite{litamin2} is not open-sourced, we compare our efficiency with \textcolor{black}{MULLS}, Suma, FAST-LIO2 and Lego-Loam. For Suma and Lego-loam, we use their own implementation on Github with the default parameters. Similar to the accuracy evaluation, all loop-closure modules of \textcolor{black}{MULLS, }Suma and Lego-Loam are completely turned off to avoid unnecessary computations. Suma uses GPUs to accelerate, while other methods only use CPU. The statistics of time consumption per LiDAR scan averaged over all KITTI training sequences are summarized in Table \ref{tab:kitti_time} Suma takes more time in our experiment than in the original paper \cite{suma2018rss} due to the difference in GPU performance. As can be seen, our method has the lowest processing time, and the next is FAST-LIO2 \cite{fastlio2}.\textcolor{black}{When compared to MULLS \cite{pan2021mulls}, we do not need feature extraction and save the time therein.} When compared to FAST-LIO2 \cite{fastlio2} and Lego-Loam \cite{shan2018lego}, we use the voxel map described in \ref{sec:voxel map} to avoid building a kd-tree in real-time \cite{fastlio2} or periodically \cite{shan2018lego}. Moreover, searching neighbor points on kd-trees has a time complexity \textcolor{black}{$O(m\log (n)) $}, \textcolor{black}{where $m$ is the dimension and} $n$ is the number of points in the dense point cloud map. In contrast, the speed of searching adjacent planes in the voxel map is near $O(1)$, which ensures efficiency. 
\begin{table}[h]
    \setlength\tabcolsep{3pt}
    \scriptsize
    \centering
    \begin{tabular}{c c c c c c}
    \hline
    Approach & Ours (full) & MULLS \cite{pan2021mulls} & Suma \cite{suma2018rss} & FAST-LIO2 \cite{fastlio2} &  Lego-Loam \cite{shan2018lego} \\
    \hline
    mean (ms) & \textbf{26.45} & 67.82 & 129.36 & 29.11 & 57.13\\
    std (ms) & \textbf{9.87} & 27.52 & 62.22 & 10.94 & 21.52\\
    \hline
    \end{tabular}
    \caption{\textcolor{black}{Comparison of Time Consumption per Scan}}
    \label{tab:kitti_time}
    \vspace{-0.5cm}
\end{table}



\subsection{
\label{ex:l515}
\textcolor{black}{Indoor Environment Test}}
\textcolor{black}{In this experiment, we test our method in indoor environments with a solid-state LiDAR Intel L515. As in experiment A, a constant velocity model is used to provide the motion prior. We use a handheld device to collect three groups of LiDAR data in laboratory and warehouse environments. All data collection routes start and end intentionally at the same place. We compare our method with SSL\_SLAM\cite{ssl2021wang}, a LiDAR SLAM approach designed for this type of LiDAR.  We run our algorithm on all datasets with the same parameters, where the maximum voxel size is 0.5m and the maximum octo-tree layer is 2. For SSL\_SLAM, we conduct the experiments using the open-sourced implementation with the default parameters.
Table \ref{tab:l515} records the end to end error and average calculation time per scan of our method and SSL\_SLAM. Compared with SSL\_SLAM, our method avoids feature extraction and considers the uncertainty of both the LiDAR point and plane map, resulting in better accuracy and efficiency than SSL\_SLAM. We also test the two methods on the dataset used by the original SSL\_SLAM paper \cite{ssl2021wang}, and the results are comparable (0.12m / 5.1ms for our method and 0.15m / 25.1ms for SSL\_SLAM \cite{ssl2021wang}). The small improvement is mainly attributed to the less challenging slow motion in the data, which was collected on a UGV moving quite stably and slowly. In contrast, the three sequences in this paper are collected on hands with much faster motions (especially rotations). Through the comparison, we see that our method is more robust to motion speed.}
\begin{table}[ht]
    \centering
    \begin{tabular}{c|c c c}
        \hline
        \multirow{2}*{Method} & \multicolumn{3}{c}{End-to-End error (m) / Avg. Comp. Time (ms)}  \\
        \cline{2-4}
        ~ & Lab1 (25.9m) & Lab2 (32.6m) & Warehouse (36.2m) \\
        \hline
        Ours (full) & \textbf{0.02/7.18} & \textbf {0.02/7.26} &\textbf {0.01/6.38} \\
        SSL\_SLAM \cite{ssl2021wang} & 0.71/27.7 & 5.80/32.9 & 0.93/28.3\\
        \hline
    \end{tabular}
    \caption{\textcolor{black}{Comparison Results on L515 Datasets}}
    \vspace{-0.2cm}
    \label{tab:l515}
\end{table}

\textcolor{black}{We further show the aggregated colored point cloud map of our method and SSL\_SLAM in the warehouse environment in Fig. 1 of the supplementary material\cite{Supplementary}. It can be seen that our colored point cloud map is more precision, while SSL\_SLAM's is more blurred, especially for the area near the return route.}



\subsection{Unstructured Environment Test}
\textcolor{black}{Both experiments in Section \ref{ex:kitti} and Section \ref{ex:l515} are carried out in structural environments. To further verify the adaptability and applicability of our method in unstructured environments, we conduct experiments in a park and a mountain environment using a non-repetitive scanning solid-state LiDAR Livox Avia. }The sensor has a built-in IMU and together they provide 10 Hz LiDAR data and 200 Hz IMU data. We use the built-in IMU to compensate the motion distortion and provide motion prior similar to FAST-LIO2 \cite{fastlio2}. In these experiments, the same computation platform and parameters \textcolor{black}{(the maximum voxel size is 3m and the maximum octo-tree layer is 3)} as in Section \ref{sec:kitti_experiment} are used despite the drastically different LiDAR scanning patterns and environment structures. \textcolor{black}{In order to validate the advantages of our map structure, we compare our results with Faster-LIO \cite{faster-lio2022ral} and FAST-LIO2 \cite{fastlio2}. The three methods use the same IKFoM framework \cite{ikfom2021he} and plane features, except for the map structure. Specifically, our method uses a probabilistic adaptive voxel map, Faster-LIO uses a Hash-indexed iVoxel map, and FAST-LIO2 uses an incremental k-d tree. For a fair comparison, 
the default 1:3 downsample used in Faster-LIO and FAST-LIO2 are turned off, so all raw points are used in all three methods. The other settings are kept default as in their open release. All results are averaged over 5 runs.}

\subsubsection{Park Environment} In this experiment, we use a handheld device to collect the LiDAR and IMU data in a large park filled with trees. In order to show the drift, the data collection route starts and ends at the same place. Besides, to avoid randomness, we collect the data of three different trajectories, of which two groups are \textcolor{black}{485m} long around the periphery of the park, and one group is \textcolor{black}{815m} long around the periphery and the interior of the park. \textcolor{black}{The mapping result of the park with the longest trajectory is shown in Fig. 2 of the supplementary material \cite{Supplementary}.} 

\subsubsection{Mountain Environment} In this experiment, we use a UAV to carry our LiDAR and face the LiDAR to the ground to collect point cloud data on a mountain. Compared to the park, the mountain environment is even less structured and more challenging. The final mapping results and some partially enlarged views can be seen in \textcolor{black}{Fig. 3 of the supplementary material}\cite{Supplementary}. We can clearly see the large planes with low uncertainty in the highway between mountains and small planes with high uncertainties on vegetation areas. 
\subsubsection{Result Analysis} 
\textcolor{black}{The end to end error and average computation time per scan of the three methods under comparison are shown in Table \ref{tab:avia_compare}. As can be seen, our method achieves much higher accuracy than both Faster-LIO and FAST-LIO2 in all three sequences, mainly due to the accurate consideration of plane uncertainty in the map where Faster-LIO and FAST-LIO2 both assume deterministic plane maps. In terms of computation efficiency, both our method and Faster-LIO use a Hash table to organize the map, so they both outperform FAST-LIO2 which uses a variant kd-tree as explained in Section \ref{ex:kitti}. When compare among our method and Faster-LIO, our method requires an additional calculation of the probability distribution of each point and the plane. While this additional calculation requires more computation time, the computed probabilistic information enables to reject further outliers as detailed in Section \ref{sec:potion-to-surface}. So the computation time of our method and Faster-LIO is more or less comparable (see Table \ref{tab:avia_compare}).} 

\begin{table}[ht]
    \vspace{-0.2cm}
    \setlength\tabcolsep{3pt}
    \centering
    \scriptsize
    \begin{tabular}{c|c c c c}
        \hline
        \multirow{2}*{Method} & \multicolumn{4}{c}{End-to-End error (m) / Avg. Comp. Time (ms)}  \\
        \cline{2-5}
        ~ & Park1 (485m) & Park2 (485m) & Park3 (815m) & Mountain (3490m)\\
        \hline
        Ours (full) & \textbf{0.03/8.1} & \textbf {0.02/8.3} & \textbf{0.01/7.7} &\textbf {6.9}/34.2 \\
        Faster-LIO \cite{faster-lio2022ral} & 0.88/9.8 & 0.99/9.8 & 0.07/9.9 & 9.8/\textbf{32.0}\\
        FAST-LIO2 \cite{fastlio2} & 1.09/12.9 & 0.76/13.4 & 0.08/12.9& 7.1/44.3\\
        \hline
    \end{tabular}
    \caption{\textcolor{black}{Comparison Results on Livox Avia Datasets}}
    \vspace{-0.3cm}
    \label{tab:avia_compare}
\end{table}

\section{Conclusion}
This paper proposes an efficient, probabilistic adaptive voxel mapping method for online LiDAR odometry. An accurate uncertainty model of the LiDAR point and plane is proposed by considering both LiDAR measurements noises and sensor pose estimation errors. The voxel map is constructed using coarse-to-fine adaptive resolution method, which is robust to different LiDAR scanning patterns and environments. This paper also shows how to implement the proposed mapping method in an iterated extended Kalman filter-based LiDAR\textcolor{black}{(-inertial)} odometry. The tests on the KITTI dataset show that our method can achieve better performance than all point cloud-based, NDT-based, and surfel-based methods. The proposed voxel mapping method not only performs well in the spinning LiDAR in structured urban environments, but also shows good performance with \textcolor{black}{solid-state LiDAR L515 in indoor environments and} non-repetitive small FoV LiDAR in unstructured environments such as parks and mountains. \textcolor{black}{A direction to improve this work is to add additional features such as edge to the voxel to expand the universality of the proposed map representation method.}



\section*{Acknowledgment}
This project is supported by DJI under the grant 200009538 and in part by SUSTech startup fund (Y01966105). The authors gratefully acknowledge Livox Technology for the equipment support during the whole work. The authors would like to thank Ambit-Geospatial and Jiarong Lin for the helps in the outdoor aerial experiment and indoor experiment.




%

\normalem
\bibliographystyle{IEEEtran}
\bibliography{paper}

\begin{thebibliography}{10}
\providecommand{\url}[1]{#1}
\csname url@samestyle\endcsname
\providecommand{\newblock}{\relax}
\providecommand{\bibinfo}[2]{#2}
\providecommand{\BIBentrySTDinterwordspacing}{\spaceskip=0pt\relax}
\providecommand{\BIBentryALTinterwordstretchfactor}{4}
\providecommand{\BIBentryALTinterwordspacing}{\spaceskip=\fontdimen2\font plus
\BIBentryALTinterwordstretchfactor\fontdimen3\font minus
  \fontdimen4\font\relax}
\providecommand{\BIBforeignlanguage}[2]{{%
\expandafter\ifx\csname l@#1\endcsname\relax
\typeout{** WARNING: IEEEtran.bst: No hyphenation pattern has been}%
\typeout{** loaded for the language `#1'. Using the pattern for}%
\typeout{** the default language instead.}%
\else
\language=\csname l@#1\endcsname
\fi
#2}}
\providecommand{\BIBdecl}{\relax}
\BIBdecl

\bibitem{liu2020low}
Z.~{Liu}, F.~{Zhang}, and X.~{Hong}, ``Low-cost retina-like robotic lidars
  based on incommensurable scanning,'' \emph{IEEE/ASME Transactions on
  Mechatronics}, pp. 1--1, 2021.

\bibitem{geiger2012we}
A.~Geiger, P.~Lenz, and R.~Urtasun, ``Are we ready for autonomous driving? the
  kitti vision benchmark suite,'' in \emph{2012 IEEE conference on computer
  vision and pattern recognition}.\hskip 1em plus 0.5em minus 0.4em\relax IEEE,
  2012, pp. 3354--3361.

\bibitem{xu2020fastlio}
W.~{Xu} and F.~{Zhang}, ``Fast-lio: A fast, robust lidar-inertial odometry
  package by tightly-coupled iterated kalman filter,'' \emph{IEEE Robotics and
  Automation Letters}, pp. 1--1, 2021.

\bibitem{okubo2009characterization}
Y.~Okubo, C.~Ye, and J.~Borenstein, ``Characterization of the hokuyo urg-04lx
  laser rangefinder for mobile robot obstacle negotiation,'' in \emph{Unmanned
  Systems Technology XI}, vol. 7332.\hskip 1em plus 0.5em minus 0.4em\relax
  International Society for Optics and Photonics, 2009, p. 733212.

\bibitem{ravi2020pothole}
R.~Ravi, A.~Habib, and D.~Bullock, ``Pothole mapping and patching quantity
  estimates using lidar-based mobile mapping systems,'' \emph{Transportation
  Research Record}, vol. 2674, no.~9, pp. 124--134, 2020.

\bibitem{zhang2014loam}
J.~Zhang and S.~Singh, ``Loam: Lidar odometry and mapping in real-time.'' in
  \emph{Robotics: Science and Systems}, vol.~2, no.~9, 2014.

\bibitem{shan2018lego}
T.~Shan and B.~Englot, ``Lego-loam: Lightweight and ground-optimized lidar
  odometry and mapping on variable terrain,'' in \emph{2018 IEEE/RSJ
  International Conference on Intelligent Robots and Systems (IROS)}.\hskip 1em
  plus 0.5em minus 0.4em\relax IEEE, 2018, pp. 4758--4765.

\bibitem{qin2020lins}
C.~Qin, H.~Ye, C.~E. Pranata, J.~Han, S.~Zhang, and M.~Liu, ``Lins: A
  lidar-inertial state estimator for robust and efficient navigation,'' in
  \emph{2020 IEEE International Conference on Robotics and Automation
  (ICRA)}.\hskip 1em plus 0.5em minus 0.4em\relax IEEE, 2020, pp. 8899--8906.

\bibitem{liosam2020shan}
T.~Shan, B.~Englot, D.~Meyers, W.~Wang, C.~Ratti, and R.~Daniela, ``Lio-sam:
  Tightly-coupled lidar inertial odometry via smoothing and mapping,'' in
  \emph{IEEE/RSJ International Conference on Intelligent Robots and Systems
  (IROS)}.\hskip 1em plus 0.5em minus 0.4em\relax IEEE, 2020, pp. 5135--5142.

\bibitem{liliom2021}
K.~Li, M.~Li, and U.~D. Hanebeck, ``Towards high-performance
  solid-state-lidar-inertial odometry and mapping,'' \emph{IEEE Robotics and
  Automation Letters}, vol.~6, no.~3, pp. 5167--5174, 2021.

\bibitem{sharp2002icp}
G.~C. Sharp, S.~W. Lee, and D.~K. Wehe, ``Icp registration using invariant
  features,'' \emph{IEEE Transactions on Pattern Analysis and Machine
  Intelligence}, vol.~24, no.~1, pp. 90--102, 2002.

\bibitem{gicp}
J.~Servos and S.~L. Waslander, ``Multi channel generalized-icp,'' in \emph{2014
  IEEE International Conference on Robotics and Automation (ICRA)}.\hskip 1em
  plus 0.5em minus 0.4em\relax IEEE, 2014, pp. 3644--3649.

\bibitem{fastlio2}
W.~Xu, Y.~Cai, D.~He, J.~Lin, and F.~Zhang, ``Fast-lio2: Fast direct
  lidar-inertial odometry,'' \emph{IEEE Transactions on Robotics}, pp. 1--21,
  2022.

\bibitem{suma2018rss}
J.~Behley and C.~Stachniss, ``Efficient surfel-based slam using 3d laser range
  data in urban environments,'' in \emph{Proc.~of Robotics: Science and
  Systems~(RSS)}, 2018.

\bibitem{chen2019suma++}
X.~Chen, A.~Milioto, E.~Palazzolo, P.~Giguere, J.~Behley, and C.~Stachniss,
  ``Suma++: Efficient lidar-based semantic slam,'' in \emph{2019 IEEE/RSJ
  International Conference on Intelligent Robots and Systems (IROS)}.\hskip 1em
  plus 0.5em minus 0.4em\relax IEEE, 2019, pp. 4530--4537.

\bibitem{wisth2021unified}
D.~Wisth, M.~Camurri, S.~Das, and M.~Fallon, ``Unified multi-modal landmark
  tracking for tightly coupled lidar-visual-inertial odometry,'' \emph{IEEE
  Robotics and Automation Letters}, vol.~6, no.~2, pp. 1004--1011, 2021.

\bibitem{marslasermap}
D.~Droeschel, M.~Schwarz, and S.~Behnke, ``Continuous mapping and localization
  for autonomous navigation in rough terrain using a 3d laser scanner,''
  \emph{Robotics and Autonomous Systems}, vol.~88, pp. 104--115, 2017.

\bibitem{quenzel2021real}
J.~Quenzel and S.~Behnke, ``Real-time multi-adaptive-resolution-surfel 6d lidar
  odometry using continuous-time trajectory optimization,'' \emph{arXiv
  preprint arXiv:2105.02010}, 2021.

\bibitem{stuckler2014multi}
J.~St{\"u}ckler and S.~Behnke, ``Multi-resolution surfel maps for efficient
  dense 3d modeling and tracking,'' \emph{Journal of Visual Communication and
  Image Representation}, vol.~25, no.~1, pp. 137--147, 2014.

\bibitem{park2017probabilistic}
C.~Park, S.~Kim, P.~Moghadam, C.~Fookes, and S.~Sridharan, ``Probabilistic
  surfel fusion for dense lidar mapping,'' in \emph{Proceedings of the IEEE
  International Conference on Computer Vision Workshops}, 2017, pp. 2418--2426.

\bibitem{elastic-continuous-tro}
C.~Park, P.~Moghadam, J.~L. Williams, S.~Kim, S.~Sridharan, and C.~Fookes,
  ``Elasticity meets continuous-time: Map-centric dense 3d lidar slam,''
  \emph{IEEE Transactions on Robotics}, 2021.

\bibitem{magnusson2007scan}
M.~Magnusson, A.~Lilienthal, and T.~Duckett, ``Scan registration for autonomous
  mining vehicles using 3d-ndt,'' \emph{Journal of Field Robotics}, vol.~24,
  no.~10, pp. 803--827, 2007.

\bibitem{2013ndtom}
J.~Saarinen, H.~Andreasson, T.~Stoyanov, J.~Ala-Luhtala, and A.~J. Lilienthal,
  ``Normal distributions transform occupancy maps: Application to large-scale
  online 3d mapping,'' in \emph{2013 ieee international conference on robotics
  and automation}.\hskip 1em plus 0.5em minus 0.4em\relax IEEE, 2013, pp.
  2233--2238.

\bibitem{litamin2}
M.~Yokozuka, K.~Koide, S.~Oishi, and A.~Banno, ``Litamin2: Ultra light
  lidar-based slam using geometric approximation applied with kl-divergence,''
  in \emph{2021 IEEE International Conference on Robotics and Automation
  (ICRA)}.\hskip 1em plus 0.5em minus 0.4em\relax IEEE, 2021, pp.
  11\,619--11\,625.

\bibitem{tatarchenko2017octree}
M.~Tatarchenko, A.~Dosovitskiy, and T.~Brox, ``Octree generating networks:
  Efficient convolutional architectures for high-resolution 3d outputs,'' in
  \emph{Proceedings of the IEEE international conference on computer vision},
  2017, pp. 2088--2096.

\bibitem{hane2017hierarchical}
C.~H{\"a}ne, S.~Tulsiani, and J.~Malik, ``Hierarchical surface prediction for
  3d object reconstruction,'' in \emph{2017 International Conference on 3D
  Vision (3DV)}.\hskip 1em plus 0.5em minus 0.4em\relax IEEE, 2017, pp.
  412--420.

\bibitem{pixel-level}
C.~Yuan, X.~Liu, X.~Hong, and F.~Zhang, ``Pixel-level extrinsic self
  calibration of high resolution lidar and camera in targetless environments,''
  \emph{IEEE Robotics and Automation Letters}, vol.~6, no.~4, pp. 7517--7524,
  2021.

\bibitem{liu2021balm}
Z.~Liu and F.~Zhang, ``Balm: Bundle adjustment for lidar mapping,'' \emph{IEEE
  Robotics and Automation Letters}, vol.~6, no.~2, pp. 3184--3191, 2021.

\bibitem{deschaud2018imls}
J.-E. Deschaud, ``Imls-slam: Scan-to-model matching based on 3d data,'' in
  \emph{2018 IEEE International Conference on Robotics and Automation
  (ICRA)}.\hskip 1em plus 0.5em minus 0.4em\relax IEEE, 2018, pp. 2480--2485.

\bibitem{pan2021mulls}
Y.~Pan, P.~Xiao, Y.~He, Z.~Shao, and Z.~Li, ``Mulls: Versatile lidar slam via
  multi-metric linear least square,'' in \emph{2021 IEEE International
  Conference on Robotics and Automation (ICRA)}.\hskip 1em plus 0.5em minus
  0.4em\relax IEEE, 2021, pp. 11\,633--11\,640.

\bibitem{ATE}
Z.~Zhang and D.~Scaramuzza, ``A tutorial on quantitative trajectory evaluation
  for visual(-inertial) odometry,'' in \emph{2018 IEEE/RSJ International
  Conference on Intelligent Robots and Systems (IROS)}, 2018, pp. 7244--7251.

\bibitem{ikfom2021he}
\BIBentryALTinterwordspacing
D.~He, W.~Xu, and F.~Zhang, ``Kalman filters on differentiable manifolds,''
  2021. [Online]. Available: \url{https://arxiv.org/abs/2102.03804}
\BIBentrySTDinterwordspacing

\bibitem{ssl2021wang}
H.~Wang, C.~Wang, and L.~Xie, ``Lightweight 3-d localization and mapping for
  solid-state lidar,'' \emph{IEEE Robotics and Automation Letters}, vol.~6,
  no.~2, pp. 1801--1807, 2021.

\bibitem{Supplementary}
C.~Yuan, W.~Xu, X.~Liu, X.~Hong, and F.~Zhang, ``Supplementary material:
  Efficient and probabilistic adaptive voxel mapping for accurate online lidar
  odometry,''
  \url{https://github.com/hku-mars/VoxelMap/blob/master/supply/VoxelMap_Supplementary_material.pdf}.

\bibitem{faster-lio2022ral}
C.~Bai, T.~Xiao, Y.~Chen, H.~Wang, F.~Zhang, and X.~Gao, ``Faster-lio:
  Lightweight tightly coupled lidar-inertial odometry using parallel sparse
  incremental voxels,'' \emph{IEEE Robotics and Automation Letters}, vol.~7,
  no.~2, pp. 4861--4868, 2022.

\end{thebibliography}

\end{document}